\documentclass{article}
\usepackage{spconf,amsmath,graphicx}
\usepackage{cite}
\usepackage{siunitx}
\usepackage{amsmath,amssymb,amsfonts}
\usepackage{graphicx}
\usepackage{textcomp}
\usepackage{booktabs}
\usepackage{xcolor}
\usepackage{bm}
\usepackage{esvect}
\usepackage[hidelinks]{hyperref} 
\def\BibTeX{{\rm B\kern-.05em{\sc i\kern-.025em b}\kern-.08em
    T\kern-.1667em\lower.7ex\hbox{E}\kern-.125emX}}
\usepackage{cite}
\usepackage{url}
\usepackage{amsmath,amssymb,amsfonts}
\usepackage{algorithm}
\usepackage{graphicx}
\usepackage{changepage}
\usepackage{algpseudocode}

\usepackage{lipsum}
\usepackage{ctable}
\usepackage{color, colortbl}

\usepackage{array, booktabs, makecell, multirow}
\definecolor{Gray}{gray}{0.9}
\usepackage{textcomp}
 
\usepackage{textcomp}
\usepackage{booktabs, makecell, multirow, tabularx}
\setcellgapes{2pt}

\usepackage{stfloats}

\usepackage{lipsum}
\usepackage{booktabs}
\newcommand\mytab[1]{\begin{tabular}[t]{@{}c@{}} #1 \end{tabular}}
\newcommand\mc[2]{\multicolumn{#1}{c}{#2}}

\usepackage{multicol}
\usepackage{subcaption}

\def\L{{\cal L}}

\begin{document}
%

\title{MCROOD: Multi-Class Radar Out-Of-Distribution Detection}
\name{Sabri Mustafa Kahya$^{\star}$ \qquad Muhammet Sami Yavuz$^{\star \dagger}$ \qquad Eckehard Steinbach$^{\star}$}
\address{  $^{\star}$Technical University of Munich,
School of Computation, Information and Technology,\vspace{-0.03cm}\\  Department of Computer Engineering, Chair of Media Technology,\vspace{-0.03cm}\\ Munich Institute of Robotics and Machine Intelligence (MIRMI) \\ $^{\dagger}$Infineon Technologies AG}


\maketitle

%
\begin{abstract}
Out-of-distribution (OOD) detection has recently received special attention due to its critical role in safely deploying modern deep learning (DL) architectures. This work proposes a reconstruction-based multi-class OOD detector that operates on radar range doppler images (RDIs). The detector aims to classify any moving object other than a person sitting, standing, or walking as OOD. We also provide a simple yet effective pre-processing technique to detect minor human body movements like breathing. The simple idea is called respiration detector (RESPD) and eases the OOD detection, especially for human sitting and standing classes. On our dataset collected by 60\si{\GHz} short-range FMCW Radar, we achieve AUROCs of 97.45\%, 92.13\%, and 96.58\% for sitting, standing, and walking classes, respectively. We perform extensive experiments and show that our method outperforms state-of-the-art (SOTA) OOD detection methods. Also, our pipeline performs 24 times faster than the second-best method and is very suitable for real-time processing.
\end{abstract}
\section{Introduction}
\label{sec:intro}

Modern deep learning architectures have shown remarkable results on several problems by making closed-world assumptions. However, this assumption is unrealistic because a model may be exposed to a sample that is out of the training distribution in an open-world setting. In this case, the model assigns the sample to one of the training classes with a high prediction probability. For safety-critical applications such as medical and autonomous driving, overconfident predictions may cause catastrophic failures. Therefore, to address this issue, in recent years, many OOD detection strategies \cite{b25,b26, b27, b28, b29, b30, b31} have been revealed. Usually, the methods consider the image or video data domain. However, the same problem also exists for various kinds of sensor data domains like radar.

Radars are robust to environmental conditions like lighting, smoke, and rain and preserve privacy. Thanks to these advantages, radars have gained interest in both industry and academia. Even though many different applications, such as gesture recognition \cite{b33}, people counting \cite{people_cnt}, and vital sign estimation \cite{vital_sign}, have been developed using radars, few of them emphasize the OOD detection problem. In this study, we aim to develop an OOD detector that operates on RDIs of 60\si{\GHz} L-shaped FMCW Radar. It tries to classify any moving object other than a person sitting, standing, or walking as OOD.

\begin{figure}[htbp]
\centerline{\includegraphics[width=0.95\linewidth]{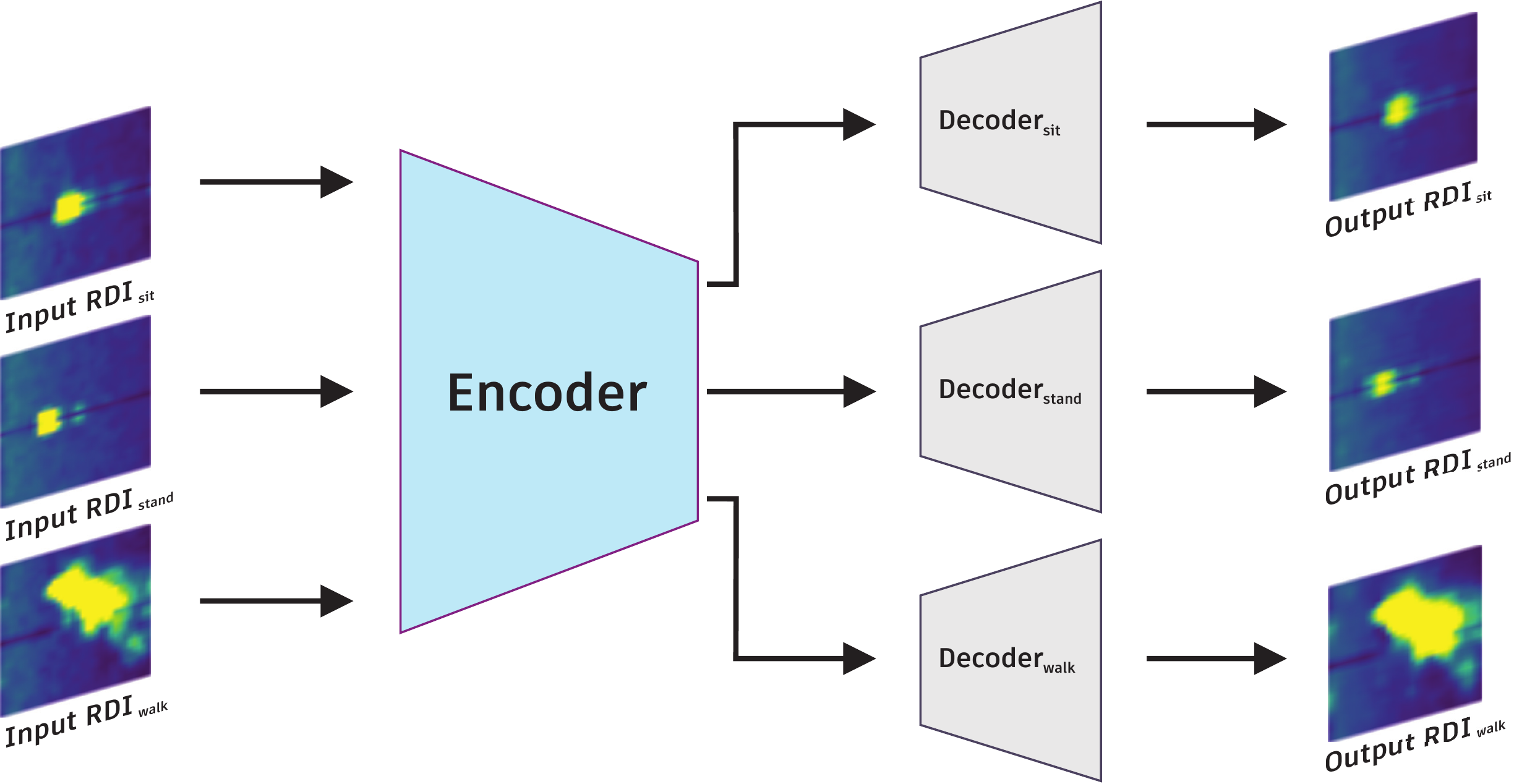}}
\caption{The overall pipeline of MCROOD.}
\label{fig:pipline}
\end{figure}

Formally, we propose a novel DL architecture (see Figure \ref{fig:pipline}) MCROOD, to perform OOD detection and a simple pre-processing technique, RESPD, to detect especially minor human body movements and to ease the OOD detection. With RESPD, instead of focusing only on a single frame's RDI, we use the combination of multiple consecutive RDIs. Our detector relies on an autoencoder-based architecture and consists of a one-encoder multi-decoder system. Each decoder corresponds to one human activity class, such as sitting, standing, and walking. Common OOD detectors use simple thresholding for detection. Based on a scoring function, a threshold is defined and used to classify the samples as ID or OOD. We use multi-thresholding. Our key contributions are as follows:
\begin{itemize}
\vspace{-0.15cm}
\item We propose a simple yet effective pre-processing idea (RESPD) to be applied to RDIs. Instead of framewise training and testing, we benefit from multiple consecutive frames at the same time for training and inference. Thus, we manage to detect human respiration and differentiate it from similar movements like moving curtains. The idea has a noticeable effect on classification performance for human sitting and standing classes.
    \vspace{-0.21cm}
    \item We also propose a novel reconstruction-based OOD detector (MCROOD) that operates on RDIs. It consists of a one-encoder multi-decoder system. The decoders correspond to the ID classes sitting, standing, and walking. We achieve AUROCs of 97.45\%, 92.13\%, and 96.58\% for sitting, standing, and walking classes, respectively, on our dataset, consisting of ID and OOD samples.
    \vspace{-0.21cm}
    \item We perform extensive evaluations on our dataset and show the superiority of MCROOD over SOTA methods in terms of common OOD detection metrics. Besides, MCROOD performs 24 times faster than the second-best method \cite{b4}. We also provide an ablation analysis to emphasize the importance of RESPD.
\end{itemize}

\section{Related Work}
\label{sec:related}
In the OOD detection field, many studies have been published using different strategies, such as post-hoc, distance-based, and outlier exposure (OE) methods.

\textbf{Post-hoc} methods aim to work on any pre-trained model. Seminal work \cite{b1} has been published as a baseline by Hendrycks and Gimpel  and relies on maximum softmax probability (MSP) scores. They mainly claim that ID samples have higher softmax scores than OOD samples. To improve \cite{b1}, a follow-up method has been revealed. ODIN \cite{b2} applies input perturbation and temperature scaling to increase the softmax scores of ID samples and to ease the detection. In energy based OOD detection study \cite{b7}, the authors apply the $logsumexp$ operator to the logit layer and get an energy value to be used for detection. ReAct\cite{b28}, as one of the recent studies, utilizes activation truncation on the penultimate layer to reduce the overconfident predictions. On the other hand, DICE \cite{dice} utilizes weight sparsification for OOD detection.

\textbf{Distance-based} methods aim to classify samples relatively far from the center of ID classes as OOD. Mahalanobis detector \cite{b4} uses intermediate feature representations and Mahalanobis distance information for detection. In a similar study \cite{b6}, the authors create uniform noise and claim that OODs are closer to noise than IDs. A recent study \cite{b31} uses a simple K-nearest-neighbor (KNN) approach to estimate the OODs. Similarly, \cite{b5, near-ood, shah} are examples of distance-based strategy.

\textbf{OE} methods expose the model to a limited number of OOD samples during training or fine-tuning. \cite{b8} uses OE with a novel loss, pushing the softmax scores of OODs to the uniform distribution. OECC \cite{b10} applies the same strategy also with a novel loss consisting of two different regularization terms. Similar studies \cite{b9, boac}, exploit OE using some OOD samples and try to generalize the knowledge to all other unseen OODs.

In the literature, there are also different strategies. As a confidence enhancement method, G-ODIN \cite{b30} proposes a new training scheme on top of \cite{b2} for better detection. GradNorm \cite{b14} emphasizes the importance of gradient information for OOD detection. \cite{b15} belongs to reconstruction-based methods and uses Mahalanobis distance information in latent space to detect OODs. 

All mentioned methods are developed and tested for the image domain. In the radar domain, there are limited DL-based methods for the OOD detection problem. In \cite{b23}, the authors compare different OOD methods on synthetically generated micro RDI data. In our work, we provide a DL-based OOD detector that works on real radar data. It consists of a one-encoder multi-decoder architecture and classifies any moving target other than a sitting, standing, or walking person as OOD. We also propose a simple pre-processing idea detecting human respiration movement for better OOD detection.
\vspace{0 cm}
\section{Radar Configuration \& Pre-processing}
We use Infineon's BGT60TR13C chipset with a 60\si{\GHz} L-Shaped FMCW radar. It consists of one transmit (Tx) and three receiver (Rx) antennas. 
The radar configuration is listed in Table \ref{tab:radar_conf}. The Tx antenna transmits $N_c$ chirp signals, and the Rx antennas receive the reflected signals. A mixture of transmitted and reflected signals produces an intermediate frequency (IF). Low-pass filtering and digitization of the IF signal result in the raw Analogue-to-Digital Converter (ADC) data. Each chirp has $N_s$ samples, so a frame's dimensions become $N_{Rx} \times N_c \times N_s$.

The IF signal's component corresponds to time within a chirp called fast time. We first apply \textbf{range FFT} on fast time to extract the range of the object. Then, we use the moving target identification (\textbf{MTI}) function on range data to remove static targets. Finally, we perform \textbf{doppler FFT} on slow time, which refers to phase across chirps and obtain the RDIs.

\begin{table}[h]
    \caption{\footnotesize FMCW Radar Configuration Parameters }
    \centering
     \footnotesize
    \begin{tabular}{@ {\extracolsep{10pt}} ccc}
    \toprule

    \centering
    Configuration name & Symbol & Value \\
    \midrule
    Number of Transmit Antennas & $N_{Tx}$  & 1  \\
    Number of Receive Antennas & $N_{Rx}$  & 3  \\
    Number of chirps per frame & $N_c$ & 64  \\
    Number of samples per chirp & $N_s$  & 128 \\
    Frame Period & $T_f$ & 50 \si{\ms}  \\
    
    Chirp to chirp time & $T_c$ & 391.55 \si{\us} \\
    Bandwidth & $B$ & 1 \si{\GHz}\\ 
    \bottomrule
    \end{tabular}

    \label{tab:radar_conf}
\end{table}
\vspace{-0.68cm}
\subsubsection*{RESPD}
RESPD is a simple pre-processing step that is applied to RDIs. Healthy adults breathe 12-20 times a minute \cite{lindh2013delmar}. To be able to detect this rhythmic movement, an about 2.5 seconds time interval is enough that corresponds to 50 consecutive frames in our radar configuration. Therefore, we pre-process the data using a sliding-window approach with a size of 50 frames. Namely, we sum all the frames within the window  and write the calculated value on the first frame in the window, then we slide the window by one frame and repeat this process until the end. RESPD has a significant effect on the classification of human sitting and standing classes.  {\color{red} }

\section{Problem Statement and MCROOD}

\begin{figure}[ht!]
    \centering
    \begin{subfigure}[b]{0.33\columnwidth}
    \includegraphics[width=\textwidth]{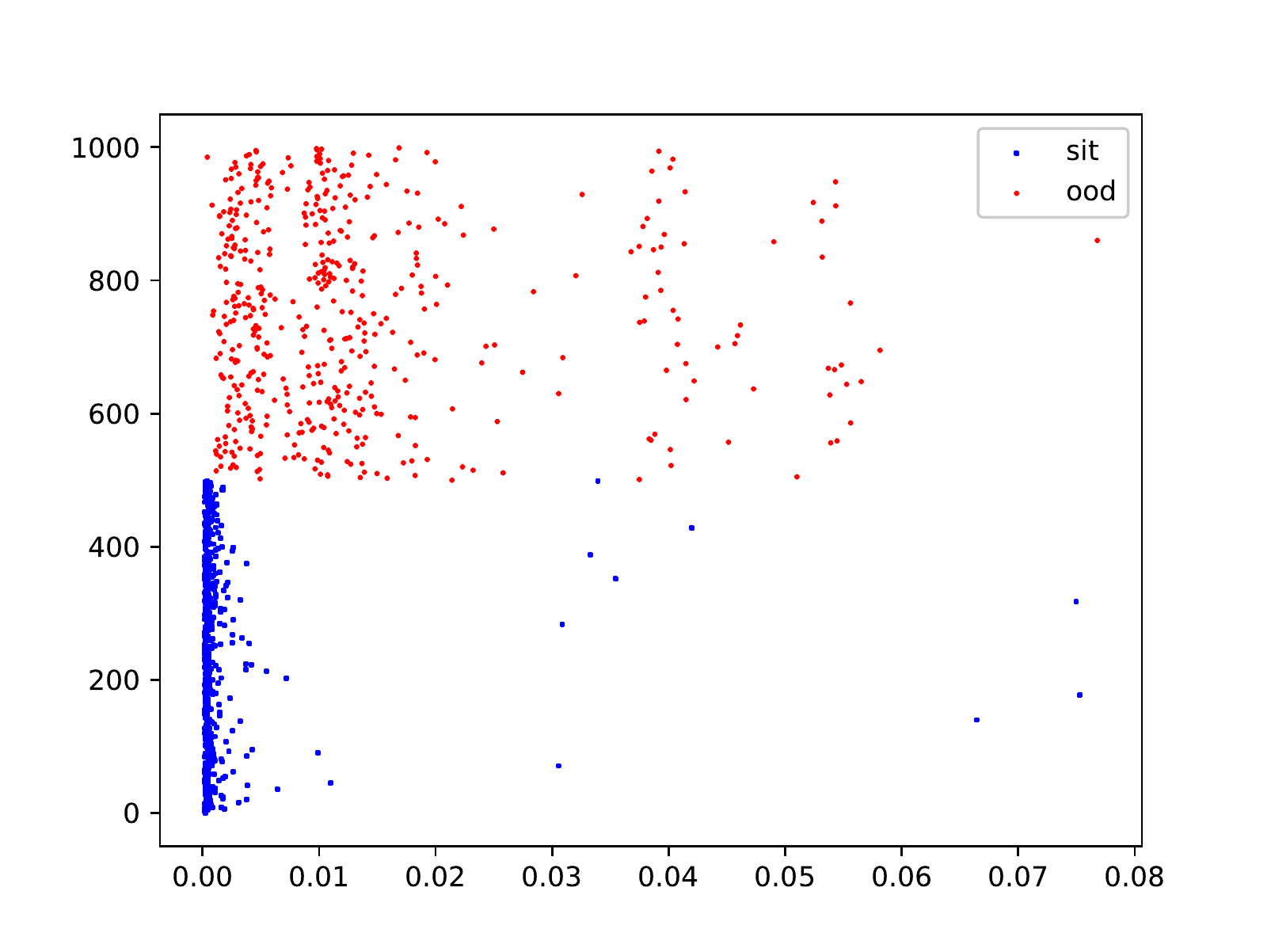}
    \caption{}
    \label{fig:sit}
    \end{subfigure}%
    \begin{subfigure}[b]{0.33\columnwidth}
    \includegraphics[width=\textwidth]{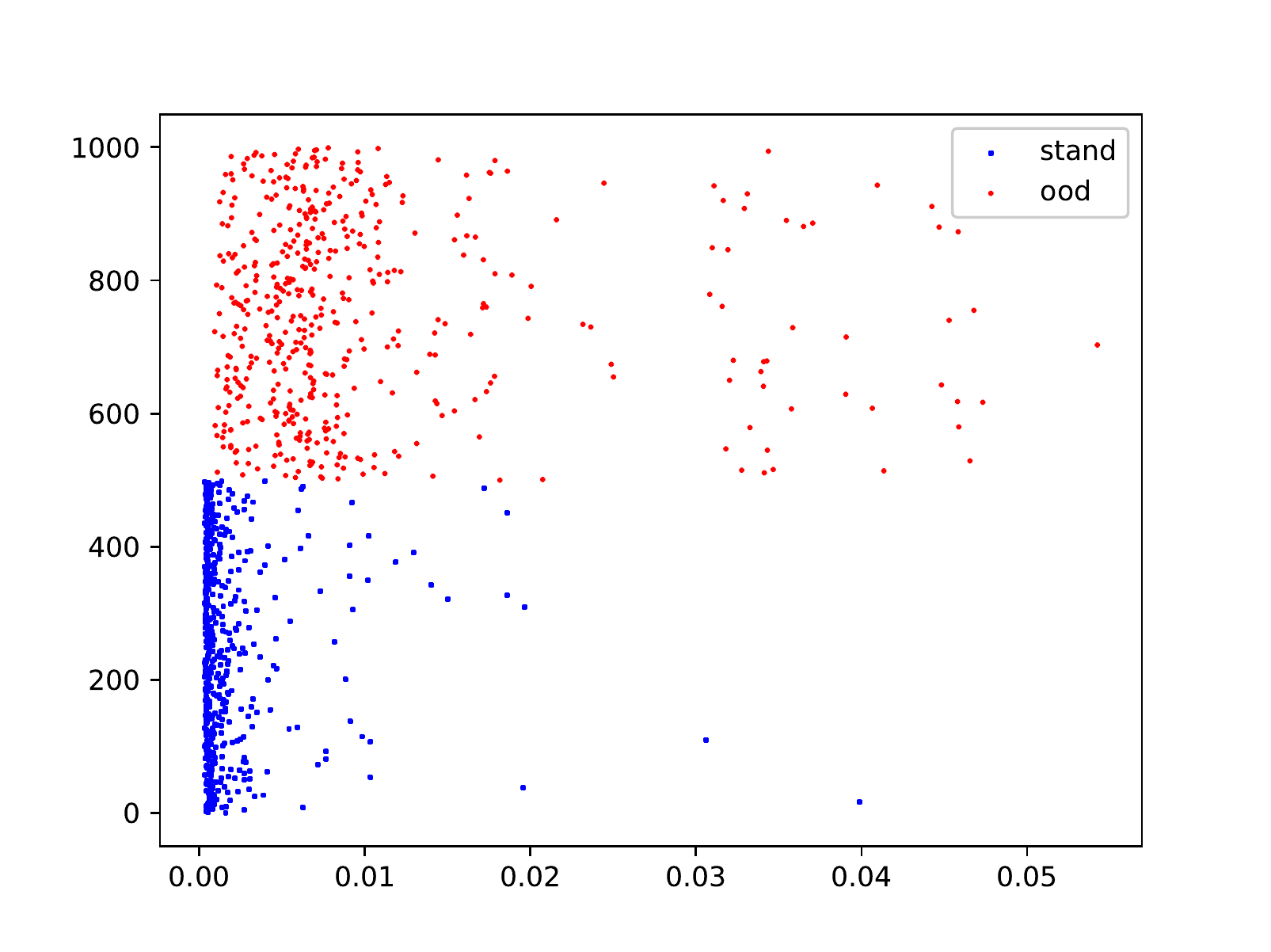}
    \caption{}
    \label{fig:stand}
    \end{subfigure}%
    \begin{subfigure}[b]{0.33\columnwidth}
    \includegraphics[width=\textwidth]{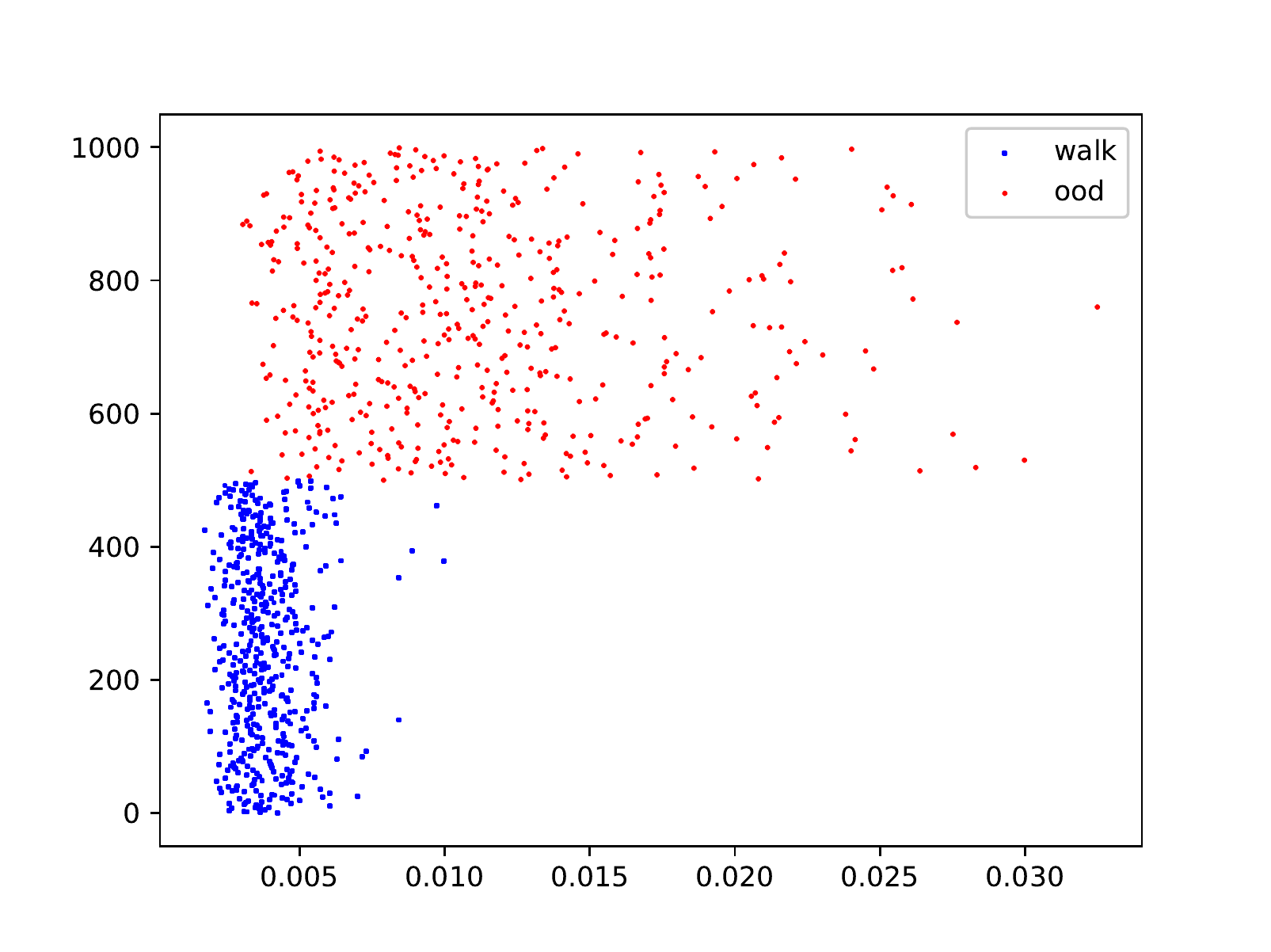}
    \caption{}
    \label{fig:walk}
    \end{subfigure}%
    \vspace{-3 mm}
     \caption{\small Reconstruction MSEs at inference time  (x-axis) of sitting (a), standing (b), walking (c) vs OODs. The y-axis represents the sample index. When IDs (blue points) aggregate in a specific region with low reconstruction errors, OODs (red points) spread out with higher error values.}
     \label{fig:scatter_plots}
     \vspace{-2mm}
\end{figure}

OOD detection is a simple binary classification task. However, in an open-world setting, a detector may be exposed to infinitely many OODs, making the binary classification task harder. We test MCROOD with a diverse set of OODs and show its robustness (see Figure \ref{fig:scatter_plots}).
\subsection{Architecture and Training} 
We use a reconstruction-based architecture. It consists of one encoder and multi (3) decoder parts. Each decoder represents one ID class (sitting, standing, and walking). The encoder consists of four main blocks. The first three blocks sequentially apply 2D convolution with 2D batch normalization and ReLu activation followed by 2D max pooling. In the blocks, for each convolutional layer, we use 16, 32, and 64 filters with 3x3 kernel size, respectively. The final block consists of flattening and dense layers followed by 1D batch normalization and outputs the latent representation. The three decoders have the same architecture. They consist of five main blocks. The first one takes the latent code and feeds it to a dense layer with a 1D batch normalization. The next three blocks sequentially apply 2D transpose convolution with 2D batch normalization and ReLu activation, followed by upsampling. The fifth block has a 2D transpose convolutional layer with sigmoid activation. In the blocks, for each transpose convolutional layer, we use 64, 32, 16, and 1 filters with 3x3 kernel size, respectively.

We perform simultaneous training for each class. Our encoder takes three inputs from three classes and produces three latent codes each time. On the other hand, as input, the decoder responsible for the sitting class only takes the latent code from sitting data. Following the same logic, the other two decoders take standing and walking data's latent codes separately. With this approach, the encoder encodes the data from three different classes while decoders only decode the class they are responsible for. As mentioned, the encoder and decoders are simultaneously trained. We use mean squared error (MSE) as our loss function.
Since there are three decoders in the network, we have three MSEs. Thus our final loss function becomes the sum of each MSE loss as in Equation \ref{eq:1}. Here $n$ is the batch size, $\textbf{X}^{(i)}_{c}$ is a data instance from class $c$, $E$ is the encoder, and $D_c$ is the class $c$'s decoder.

\begin{equation}
\footnotesize
\begin{aligned}
\L& = \frac{1}{n} \sum_{i=1}^n (\textbf{X}^{(i)}_{sit} - D_{sit}(E(\textbf{X}^{(i)}_{sit})))\\
&+\frac{1}{n}\sum_{i=1}^n (\textbf{X}^{(i)}_{stand} - D_{stand}(E(\textbf{X}^{(i)}_{stand})))\\ &+\frac{1}{n}\sum_{i=1}^n (\textbf{X}^{(i)}_{walk} - D_{walk}(E(\textbf{X}^{(i)}_{walk})))
\end{aligned}
\label{eq:1}
\end{equation}

\subsection{OOD Detection}
Reconstruction-based OOD detection methods detect the OODs based on the difference between the input and reconstructed output. The detector classifies the sample as OOD if the difference is more than a threshold. In MCROOD, we expect to see more reconstruction errors for OODs since we train our network using only IDs.
Since we have a multi-decoder system, we perform multi-thresholding. During inference, we feed the encoder with the same input (X) from its three input gates. The encoder encodes and gives the latent code to each decoder separately. After the reconstruction, we separately evaluate the errors between the input and outputs by reconstruction MSE and then perform the multi-thresholding. The network classifies the sample as OOD if all reconstruction errors exceed their corresponding thresholds; otherwise, ID.

\vspace{-0.4cm}
\section{Experiments}

We perform our experiments with a processing unit of NVIDIA GeForce RTX 3070 GPU, Intel Core i7-11800H CPU, and 32GB DDR4 module of RAM.

\begin{table*}[ht]
\centering
\footnotesize
\caption{Main Results. OOD detection performance comparison with other popular methods. All values are shown in percentages. $\uparrow$ indicates that higher values are better, while $\downarrow$ indicates that lower values are better.}
\label{Results}
\setlength\tabcolsep{0pt}
\begin{tabular*}{\textwidth}{@{\extracolsep{\fill}}c|ccccccccccccc}
    \toprule
    & 
    & \mc{4}{\mytab{Sit}} & \mc{4}{\mytab{Stand}}
    & \mc{4}{\mytab{Walk}} 
   \\
    \cmidrule{3-6} \cmidrule{7-10} \cmidrule{11-14}
\centering
    Architecture & Methods 
    & AUROC
    & AUPR 
    & FPR95 
    & FPR80 
    & AUROC 
    & AUPR 
    & FPR95 
    & FPR80 
    & AUROC 
    & AUPR 
    & FPR95
    & FPR80 
    \\
    &  
    &  $\uparrow$
    &  $\uparrow$
    &  $\downarrow$
    &  $\downarrow$
    &  $\uparrow$
    &  $\uparrow$
    &  $\downarrow$
    &  $\downarrow$
    &  $\uparrow$
    &  $\uparrow$
    &  $\downarrow$
    &  $\downarrow$
    \\   
      
    \midrule
     & ODIN \cite{b2} &69.51&67.62&66.64&47.43&56.11&56.50&87.19&72.08&79.11&71.99&47.27&36.82 \\

    & MSP \cite{b1} &50.67&54.12&91.82&77.85&39.89&42.25&94.34&83.73&89.71&85.31&31.72&20.46   \\

    & ENERGY \cite{b7} &50.89&53.52&85.20&75.06&39.73&41.59&88.93&81.85&87.78&82.64&35.42&23.45   \\

    RESNET-34 \cite{resnet}& MAHA \cite{b4} & 96.50&96.95&10.24&5.50&86.55&81.62&46.87&20.06&66.73&53.73&73.40&58.55 \\
    &FSSD \cite{b6} &40.33&45.67&83.67&65.82&47.65&43.10&66.20&59.07&\textbf{96.67}&\textbf{95.09}&18.46&\textbf{2.82} \\
    &OE\cite{b8} &49.73&55.40&94.89&80.30&49.72&48.97&95.30&79.45&49.99&43.55&94.97&80.04  \\
    \midrule
    \midrule
    \addlinespace
    
    & ODIN\cite{b2} &57.13&53.78&71.42&56.66&53.14&47.11&80.59&64.61&94.63&92.90&24.06&9.36 \\

    & MSP \cite{b1} & 47.31&49.25&90.02&73.54&40.67&41.10&92.88&81.58&95.42&93.08&19.95&7.96 \\

    & ENERGY\cite{b7} & 49.23&49.64&80.48&69.19&41.39&40.89&88.50&76.66&94.89&92.82&21.28&8.44  \\

    RESNET-50 \cite{resnet}& MAHA \cite{b4} & 97.18&97.60&\textbf{9.95}&4.48&90.35&85.13&\textbf{33.11}&12.64&82.81&68.44&53.21&23.83 \\
    &FSSD\cite{b6}&6.62&35.38&99.90&99.76&21.11&34.05&99.58&94.74&92.53&86.96&19.95&13.21 \\
    &OE\cite{b8}&38.23&46.90&99.23&90.22&26.27&35.85&96.50&91.37&94.33&91.23&15.29&11.76  \\
    \addlinespace
    \midrule
    \midrule
    & ODIN\cite{b2} &55.00&53.84&84.36&67.01&54.66&48.54&81.61&68.98&94.14&92.05&20.29&13.83  \\

    & MSP \cite{b1} & 45.31&49.07&94.69&82.84&45.27&43.40&93.58&80.32&93.44&87.43&18.83&13.41  \\

    & ENERGY \cite{b7} &47.46&49.93&94.05&75.85&47.59&44.62&91.31&74.87&94.51&92.40&18.96&13.36  \\
    \centering
    RESNET-101 \cite{resnet}& MAHA \cite{b4} &94.21&94.94&21.37&8.88&89.91&85.15&33.62&12.44&72.82&57.57&67.71&41.84  \\
    &FSSD\cite{b6}&12.74&37.38&99.80&98.71&23.73&35.38&99.66&97.99&90.59&89.10&49.53&16.35 \\
    &OE\cite{b8}&39.22&45.34&98.30&75.27&44.37&42.08&95.00&69.83&95.98&94.05&14.60&8.08 \\
\addlinespace
    \midrule
    \midrule
    
Ours &MCROOD & \textbf{97.45}&\textbf{98.40}&12.93&\textbf{0.35}&\textbf{92.13}&\textbf{93.51}&50.94&\textbf{9.90}&96.58&94.82&\textbf{13.91}&5.24 \\

    
    \bottomrule

\end{tabular*}
\end{table*}

\vspace{-0.4cm}
\subsection{Dataset and Evaluation Metrics}

Data is collected using Infineon’s BGT60TR13C 60\si{\GHz} FMCW radar sensor with four individuals in 16 house, school, and office rooms (10 for training and six for inference). It consists of ID and OOD samples. IDs include human walking, sitting, and standing classes with changing ranges and distances from 1 to 5m. OOD samples include a table fan, a remote-controlled (RC) toy car, swinging plants, swinging blinds, swinging curtains, swinging laundry, a swinging paper bag, a vacuum cleaner, and a robot vacuum cleaner. We have 111416 ID frames in our training set. During inference, we used 47210 ID and 16050 OOD frames. 

We chose commonly used evaluation metrics AUROC, AUPR, FPR95, and FPR80. \textbf{AUROC} is the area under the receiver operating characteristic (ROC) curve. \textbf{AUPR} is the area under the precision-recall curve. \textbf{FPR95} is the false positive rate (FPR) when the true positive rate (TPR) is 95\%. \textbf{FPR80} is the FPR when TPR is 80\%.

We compare MCROOD with six SOTA methods in terms of common OOD metrics. For a detailed comparison, we separately train ResNet34, ResNet50, and ResNet101 backbones \cite{resnet} with our three class ID data in a multi-class classification manner. During training, no OOD data are exposed to the architectures. Afterward, we use the pre-trained models to apply the SOTA methods and evaluate their performance in terms of significant metrics. Table \ref{Results} shows the superiority of MCROOD over SOTA approaches. A short demo video of our system is available at the link\footnote{\href{https://figshare.com/s/b6562822d95f4d5c5c47} {https://figshare.com/s/b6562822d95f4d5c5c47}}.

\vspace{-0.28cm}
\subsection{Ablation}
We also perform ablation experiments to show the impact of RESPD. For this experiment, we only choose the best two methods, namely MCROOD and MAHA \cite{b4}, and perform a framewise evaluation on the same models without RESPD. Instead of using consecutive 50 frames, we evaluate each frame separately to perform OOD detection. Table \ref{table:ablation} shows the effectiveness of RESPD, especially for the ID samples of sitting and standing classes. On its nature, the walking class produces macro movements, so RESPD has limited impact on it while distinguishing IDs from OODs. When all test samples are evaluated, MCROOD and MAHA \cite{b4} finish their inference in 129 and 3121 seconds, respectively.

\begin{table}[ht!]
\footnotesize
\caption{Ablation Study. All values are shown in percentages. $\uparrow$ indicates that higher values are better.}   
\centering
\begin{tabular}{@{\extracolsep{\fill}}cccc}
\toprule   
{} &{}&\multicolumn{1}{c}{AUROC $\uparrow$ }  \\

 \cmidrule{2-4} 
\centering 
  Methods & Sit  & Stand & Walk \\ 
\midrule
 MAHA \cite{b4}   &14.93 &24.68&85.85   \\
MCROOD &12.69&20.73&89.43\\ 
MAHA+RESPD &97.18&90.35&82.81

\\ 
MCROOD+RESPD &\textbf{97.45}&\textbf{92.13}&\textbf{96.58} 

\\
\bottomrule
\end{tabular}
\label{table:ablation} 
\end{table}

\vspace{0.1cm}
\section{Conclusion}
In this study, we developed an OOD detection pipeline to be performed on 60\si{\GHz} FMCW radar data. Our detector MCROOD has a novel reconstruction-based one-encoder multi-decoder architecture; due to its modular nature, it can work with any number of in-class activities. We used our detector to classify any moving object which may appear in indoor environments other than a sitting, standing, or walking human as OOD. We also provided RESPD as a simple yet effective pre-processing idea. With RESPD, we aimed to detect minor human body movements like breathing. MCROOD with RESPD reached AUROCs of 97.45\%, 92.13\%, and 96.58\% for sitting, standing, and walking classes, respectively. Our experiments show that MCROOD outperforms popular SOTA methods. Besides, compared with the best second method, MCROOD has a 24 times faster processing time and is very suitable for real-time evaluation.

\vfill \pagebreak


\bibliographystyle{ieeetr}


\end{document}